\documentclass[pdflatex,sn-vancouver,Numbered]{sn-jnl}

\usepackage{graphicx}%
\usepackage{multirow}%
\usepackage{amsmath,amssymb,amsfonts}%
\usepackage{amsthm}%
\usepackage{mathrsfs}%
\usepackage[title]{appendix}%
\usepackage{textcomp}%
\usepackage{manyfoot}%
\usepackage{booktabs}%
\usepackage{algorithm}%
\usepackage{algorithmicx}%
\usepackage{algpseudocode}%
\usepackage{listings}%

\usepackage{threeparttable}
\usepackage{array}
\usepackage{booktabs}
\usepackage{pifont} 
\usepackage[table]{xcolor} 
\usepackage{setspace}

\newcommand{\cmark}{\textcolor{green}{\ding{51}}} 
\newcommand{\xmark}{\textcolor{red}{\ding{55}}} 
\newcommand{\ift}{\textcolor{orange}{IFT}} 
\newcommand{\tfp}{\textcolor{blue}{TFP}} 
\newcommand{\unknown}{\textcolor{gray}{Unknown}} 
\newcommand{\np}{\textcolor{gray}{N/P}} 
\newcommand{\na}{\textcolor{gray}{N/A}} 
\newcommand{\mentalhealth}{\textcolor{orange}{Mental Health}} 
\newcommand{\lang}{\textcolor{blue}{Language}}

\begin{document}

\title[Article Title]{Large Language Models in Mental Health Care:\\ A Scoping Review}


\author[1,2]{\fnm{Yining} \sur{Hua}}

\author[3]{\fnm{Fenglin} \sur{Liu}}

\author[4]{\fnm{Kailai} \sur{Yang}}

\author[5]{\fnm{Zehan} \sur{Li}}

\author[6]{\fnm{Hongbin} \sur{Na}}

\author[7,8]{\fnm{Yi-han} \sur{Sheu}}

\author[9]{\fnm{Peilin} \sur{Zhou}}

\author[10]{\fnm{Lauren V.} \sur{Moran}}

\author[4,11]{\fnm{Sophia} \sur{Ananiadou}}

\author[3]{\fnm{David A.} \sur{Clifton}}

\author[1]{\fnm{Andrew} \sur{Beam}}

\author*[2,8]{\fnm{John} \sur{Torous}}\email{jtorous@gmail.com}

\affil[1]{\orgdiv{Department of Epidemiology}, \orgname{Harvard T.H. Chan School of Public Health}, \orgaddress{\state{Massachusetts}, \country{USA}}}

\affil[2]{\orgdiv{Department of Psychiatry}, \orgname{Beth Israel Deaconess Medical Center}, \orgaddress{\state{Massachusetts}, \country{USA}}}

\affil[3]{\orgdiv{Institute of Biomedical Engineering}, \orgname{University of Oxford}, \orgaddress{\state{Oxford}, \country{UK}}}

\affil[4]{\orgdiv{Department of Computer Science}, \orgname{The University of Manchester}, \orgaddress{\state{Manchester}, \country{UK}}}

\affil[5]{\orgdiv{School of Biomedical Informatics}, \orgname{University of Texas Health Science at Houston}, \orgaddress{\state{Houston}, \country{USA}}}

\affil[6]{\orgdiv{Australian Artificial Intelligence Institute}, \orgname{University of Technology Sydney}, \orgaddress{\state{New South Wales}, \country{AUS}}}

\affil[7]{\orgdiv{Center for Precision Psychiatry}, \orgname{Massachusetts General Hospital}, \orgaddress{\state{Massachusetts}, \country{USA}}}

\affil[8]{\orgdiv{Department of Psychiatry}, \orgname{Harvard Medical School}, \orgaddress{\state{Massachusetts}, \country{USA}}}

\affil[9]{\orgdiv{Data Science and Analytics Thrust}, \orgname{Hong Kong University of Science and Technology}, \orgaddress{\state{Guangzhou}, \country{CHN}}}

\affil[10]{\orgdiv{Division of Psychotic Disorders}, \orgname{McLean Hospital}, \orgaddress{\state{Massachusetts}, \country{USA}}}

\affil[11]{\orgname{The Alan Turing Institute}, \orgaddress{\state{London}, \country{UK}}}




\abstract{\textbf{Objectieve:} This review aims to deliver a comprehensive analysis of Large Language Models (LLMs) utilization in mental health care, evaluating their effectiveness, identifying challenges, and exploring their potential for future application.

\textbf{Materials and Methods:} A systematic search was performed across multiple databases including PubMed, Web of Science, Google Scholar, arXiv, medRxiv, and PsyArXiv in November 2023. The review includes all types of original research, regardless of peer-review status, published or disseminated between October 1, 2019, and December 2, 2023. Studies were included without language restrictions if they employed LLMs developed after T5 and directly investigated research questions within mental health care settings.

\textbf{Results:} Out of an initial 313 articles, 34 were selected based on their relevance to LLMs applications in mental health care and the rigor of their reported outcomes. The review identified various LLMs applications in mental health care, including diagnostics, therapy, and enhancing patient engagement. Key challenges highlighted were related to data availability and reliability, the nuanced handling of mental states, and effective evaluation methods. While LLMs showed promise in improving accuracy and accessibility, significant gaps in clinical applicability and ethical considerations were noted.

\textbf{Conclusion:} LLMs hold substantial promise for enhancing mental health care. For their full potential to be realized, emphasis must be placed on developing robust datasets, development and evaluation frameworks, ethical guidelines, and interdisciplinary collaborations to address current limitations.
}




\maketitle

\section{Introduction}\label{sec:introduction}
Mental health is a paramount component of modern public health. 
Statistics from the US National Institute of Mental Health (NIMH) show that 22.8\% of U.S. adults experienced mental illness of any type in 2021 \cite{NAMI2023}. Globally, mental health disorders account for 30\% of the non-fatal disease burden, highlighting their prevalence as a major cause of disability as published by the World Health Organization (WHO) \cite{Marquez2016Mental}. Moreover, the WHO estimates that depression and anxiety disorders cost the global economy \$1 trillion per year in lost productivity \cite{WHO2023SpecialInitiative}. These data emphasize the critical need for innovations in the prevention and management of mental health conditions, both to alleviate individual suffering and to reduce the substantial societal and economic burdens associated with mental health issues.

A large portion of mental health issues and their management take place within the realm of natural language. This encompasses a range of elements, including the evaluation of symptoms and signs associated with mental health disorders and various forms of interventions, such as talk therapies. In this context, the wealth of information embedded in recorded textual expressions and interactions serves as a crucial resource, enhancing research and practical applications in mental health. 

Accordingly, Natural Language Processing (NLP), a branch of computer science that enables computers to process unstructured text (natural language) information in meaningful ways, has shown promise as a tool to aid mental health-related tasks. This includes identifying specific mental conditions\cite{Zhang2022}, supporting interventions \cite{Malgaroli2023}, and building emotion-support chatbots \cite{Li2023}. These tasks utilize data from various sources, such as clinical data \cite{irving2021using, levis2021natural} and social media data\cite{socialmediareview, coppersmith2014quantifying, li2022tracking, wu2023exploring}.

Large Language Models (LLMs), one of the most recent advances in NLP, have further expanded the potential for innovative mental health care\cite{Heerden, ji2023rethinking}. As a type of Artificial Intelligence (AI) that understands and generates human-like fluent texts, LLMs offer many promising applications for mental health. For example, their ability to efficiently retrieve and summarize information from large volumes of unstructured text contained in diverse sources (e.g., electronic health records, mobile device interactions, social media platforms, etc.) can provide insights into patient behaviors and experiences for clinicians. This may help with early intervention strategies and tailored treatment plans. In addition, through conversational interactions, they have the potential to offer a relatable and accessible means for individuals to articulate their emotional states and personal experiences\cite{2}. This aids in both user self-expression and also potentially enhances clinical therapeutic processes if they can be successfully validated and integrated into the clinical workflows.

While the interest in deploying LLMs to mental health care is increasing, to our knowledge, there are no reviews on this topic. This study aims to bridge this gap, presenting the first comprehensive review of LLM applications in mental health care. To differentiate LLMs from smaller-sized language models, we follow an established review for general domain LLMs \cite{zhao2023survey} to focus exclusively on model architectures developed since the introduction of T5 \cite{t5} in 2019. This period, spanning the last four years, marks a significant phase in the evolution of LLMs, offering a rich context to explore their influence and advancements in the field of mental health care. Through this study, we aim to explore:

\begin{enumerate}
    \item Datasets, models, and training techniques;
    \item Mental health applications, conditions, and validation measures;
    \item Ethical, privacy, safety, and other challenges.
    \item Gaps between tools currently available and clinical practicality.
\end{enumerate}


\subsection{Background on Natural Language Processing and Large Language Models}
Natural Language Processing (NLP) emerged in the late 1940s with machine translation systems \cite{jones1994natural}. Initially, the focus was on rule-based systems that required explicit coding of language rules\cite{eliza, schank1972conceptual}, which limited their scope and adaptability to handle the complexity and variability of natural languages. Statistical Language Models arose in the 1990s and they modeled word sequences using statistical methods\cite{beeferman1999statistical}. These models were typically hindered by issues such as data sparsity. Neural language models, proposed in the 2000s, utilized neural network-inspired architectures (e.g., multi-layer perceptrons and recurrent neural networks) to refine the prediction of word sequences\cite{neuralmodels}. This era witnessed the emergence of distributed word representations, notably through the word2vec algorithm\cite{word2vec}. which shifted the focus towards more effective feature learning for varied NLP tasks. Pre-trained language models like ELMo\cite{elmo}, BERT\cite{bert}, T5\cite{t5}, and GPT \cite{gpt1} were introduced in the late 2010s. These models leveraged advanced neural architectures, including bidirectional LSTM \cite{bi-lstm} and Transformers\cite{transformer}, revolutionizing NLP with their `pre-training and fine-tuning' paradigm. 
Starting from around 2020, NLP gradually moved toward the 'large language model (LLM)' era. LLMs can be analogized to sophisticated, pre-built statistical models but are massive in their scales and can offer much more in terms of versatility and depth. 


Typical LLMs include GPT-3\cite{gpt3}, GPT-3.5\cite{gpt35}, GPT-4 \cite{gpt-4} (ChatGPT is an application developed based on GPT-3.5 and GPT-4), LLaMA\cite{llama}, LLaMA-2 \cite{llama2} and PaLM\cite{palm}. Thanks to their impressive performance in understanding and generating fluent human language, natural language "prompts" are proposed to interact with and instruct these LLMs with their internal weights (i.e. model parameters) fixed\cite{promptlearning}. This strategy is generally referred to as "prompt-tuning," a paradigm distinct from "fine-tuning," where adjusting model weights requires training with additional data.

Today, due to differing policies from developers, LLMs are divided into two categories: open-source and non-opensource. Open-source LLMs, e.g., LLaMA, can be deployed in local environments, allowing for customization with specific datasets. This local deployment ensures data privacy is maintained, as it obviates the necessity for data to be uploaded to cloud-based services, which is crucial in sensitive fields like mental health. In contrast, non-open-source LLMs like GPT-4 and PaLM do not share their model architectures or training details. These models are typically hosted by third-party organizations and can be accessed on their websites or through Application Programming Interfaces (APIs). While offering robust language processing capabilities, they require different data privacy and control considerations.

\subsection{Potential Applications in Mental Health Care}
LLMs, effective in processing large volumes of natural language texts and simulating human-like interactions, have the potential to assist with a wide range of tasks in mental health care. These include interpreting and predicting behavioral patterns, identifying psychological stressors, and providing emotional support. With adequate regulatory, ethical, and privacy safeguards, they may also serve as supplementary roles to support clinically-oriented tasks. This could involve aiding in preliminary assessments for diagnostic processes, facilitating the management of psychiatric disorders, and enhancing therapeutic interventions, such as by helping to improve treatment adherence. 

\section{Methods}\label{sec:methods}
\subsection{Search Strategy and Selection Criteria}\label{sec:search}
\begin{figure}[th!]
    \centering
    \includegraphics[width=0.65\linewidth]{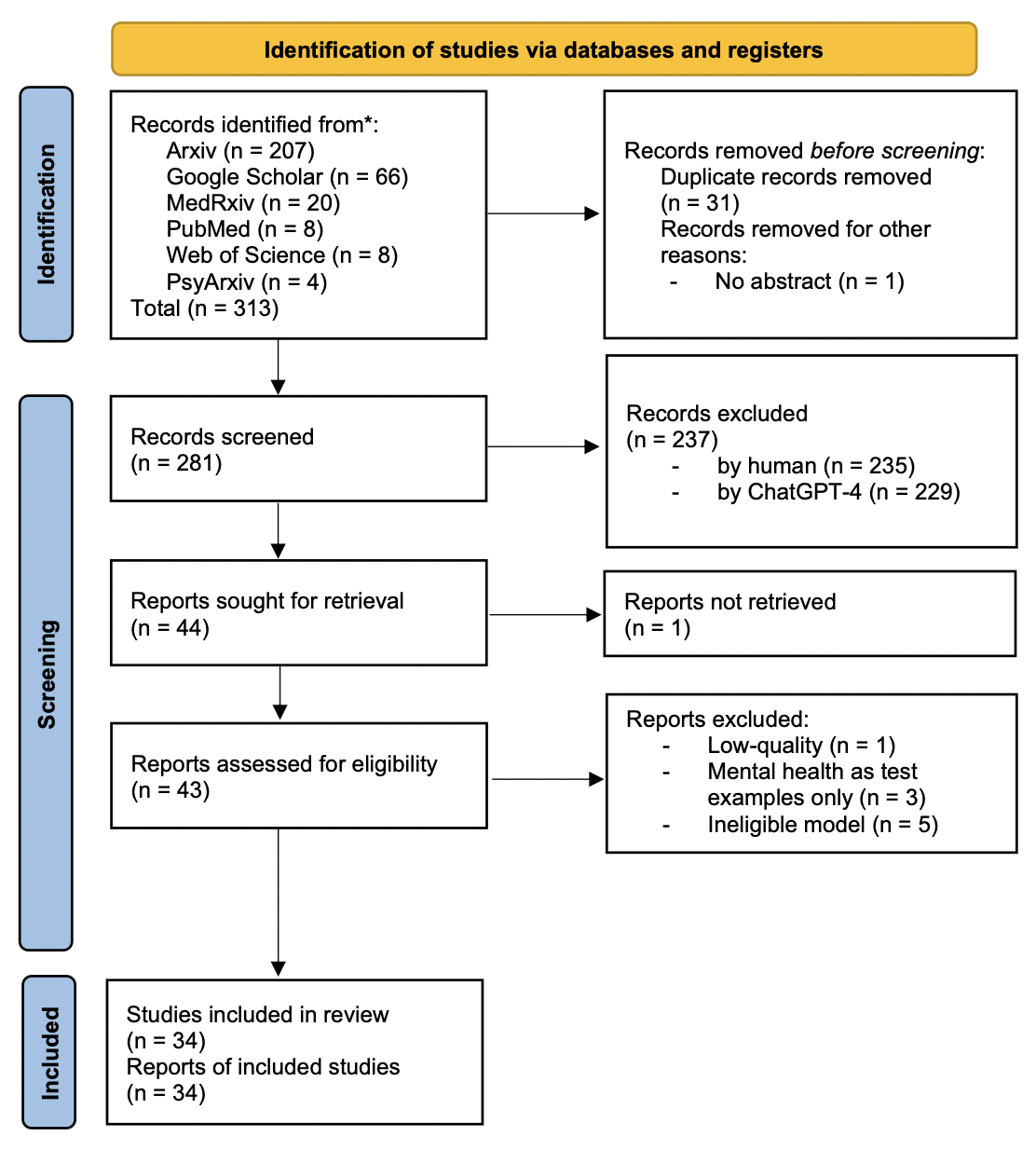}
    \caption{Screening prompt for ChatGPT-4.}
    \label{fig:prisma}
\end{figure}

In conducting this review, our approach adhered to the Preferred Reporting Items for Systematic Reviews and Meta-Analysis (PRISMA) guidelines (2020 version) \cite{prisma} to ensure a transparent and rigorous process. The PRISMA diagram is shown in Figure \ref{fig:prisma}.

We established specific criteria for paper inclusion: (1) The study must employ at least one LLM published after T5, as defined in\cite{zhao2023survey}, and (2) it should directly address research questions of these models in mental health care settings. This model scope was chosen to ensure that our review concentrates on the latest advancements in the field, deliberately excluding early pre-trained language models such as BERT and GPT-2, which have already received extensive coverage in prior literature.

Our initial search on PubMed highlighted a limited number of published studies in this domain. Recognizing the fast-evolving nature of LLMs, we extended our scope beyond traditional peer-reviewed literature. This inclusion of both peer-reviewed and non-peer-reviewed studies, such as preprints, is essential to capture the latest advancements in LLMs, which are rapidly advancing and often reported in non-traditional formats. We considered all forms of original research, including long and short articles and case reports, published or disseminated between October 1, 2019, and Dec 2, 2023, without any language restrictions.

We conducted comprehensive searches across several databases and registries, including ArXiv, MedRxiv, the ACM Digital Library, PubMed, Web of Science, and Google Scholar. The search strategy used the logical combination of keywords "Large Language Model" AND "mental OR psychiatry OR psychology." Whenever possible, we restricted the search scope to titles and abstracts. In databases without this functionality, we extended the search scope to the full text of the papers.

Further details regarding the search process and specific search queries can be found in Appendix \ref{sec:prompt_design}. For the Google Scholar registry, we employed the same set of keywords and reviewed the first 100 studies, ordered by relevance, that met our inclusion criteria. 

\subsection{Screening with GPT-4}\label{subsec:screening}
After removing duplicated articles and those without abstracts, we had 281 papers left for initial screening. Recognizing that GPT-4 has been shown to be adept in assisting with article screening and exhibits performance comparable to humans\cite{gpt4review2, gpt4review1}, we used GPT-4 as a secondary reviewer for this process. Prior to its use, we tested different prompts to optimize GPT-4's screening capability (\ref{sec:prompt_design}).

Both YH and GPT-4 independently screened the article titles and abstracts to determine their eligibility for inclusion in the study. Each had three response options: 1 for inclusion, 0 for exclusion, and 2 for uncertainty. Any disagreements identified were subsequently resolved through discussion with the rest of the review team (YH, KY, ZL, and FL). To quantitatively evaluate the agreement level between the human reviewer (YH) and the AI (GPT-4), Cohen's Kappa \cite{McHugh2012} was calculated to be approximately 0.9024. This high score indicates a strong agreement. It was observed that GPT-4 tended to be more inclusive, often categorizing more articles as related to mental health care than the human reviewer. Nonetheless, implementing the option of uncertainty, while slightly reducing the kappa score, was crucial in ensuring comprehensive inclusion of relevant articles, thereby balancing thoroughness with precision.

\section{Findings}
Forty-three articles remained for full-text review. YH, KY, ZL, and FL reviewed all articles and removed nine articles due to low quality (n = 1), only having mental health as test cases (n = 3), or ineligible model size (n = 5). During the review, studies are categorized based on their research questions and intentions:
\begin{enumerate}
    \item Dataset and Benchmark: These studies use standardized tests or benchmark datasets to evaluate and compare the performance of different methods, systems, or models under controlled conditions. Some of them curate or compile new datasets.
    \item Methods Development: These studies propose new LLMs or introduce methods, such as fine-tuning and prompting, to adapt and improve existing LLMs for mental health care.
    \item Specific Applications: These studies employ LLMs to evaluate their performance in mental health-related tasks in real-world applications. They may include the evaluation of LLMs on certain tasks (inference-only).
    \item Ethics, Privacy, and Safety consideration: These studies examine the potential risks, ethical dilemmas, and privacy concerns associated with deploying LLMs in sensitive mental health contexts and propose frameworks or guidelines to mitigate these issues.
\end{enumerate}
Thirty-four articles met our criteria and were included in subsequent analyses. To maintain a clear focus and ensure a thorough analysis to answer the application-focused study questions, we summarized "Dataset and Benchmark" studies separately.

\begin{figure}[th!]
    \centering
    \includegraphics[width=0.85\linewidth]{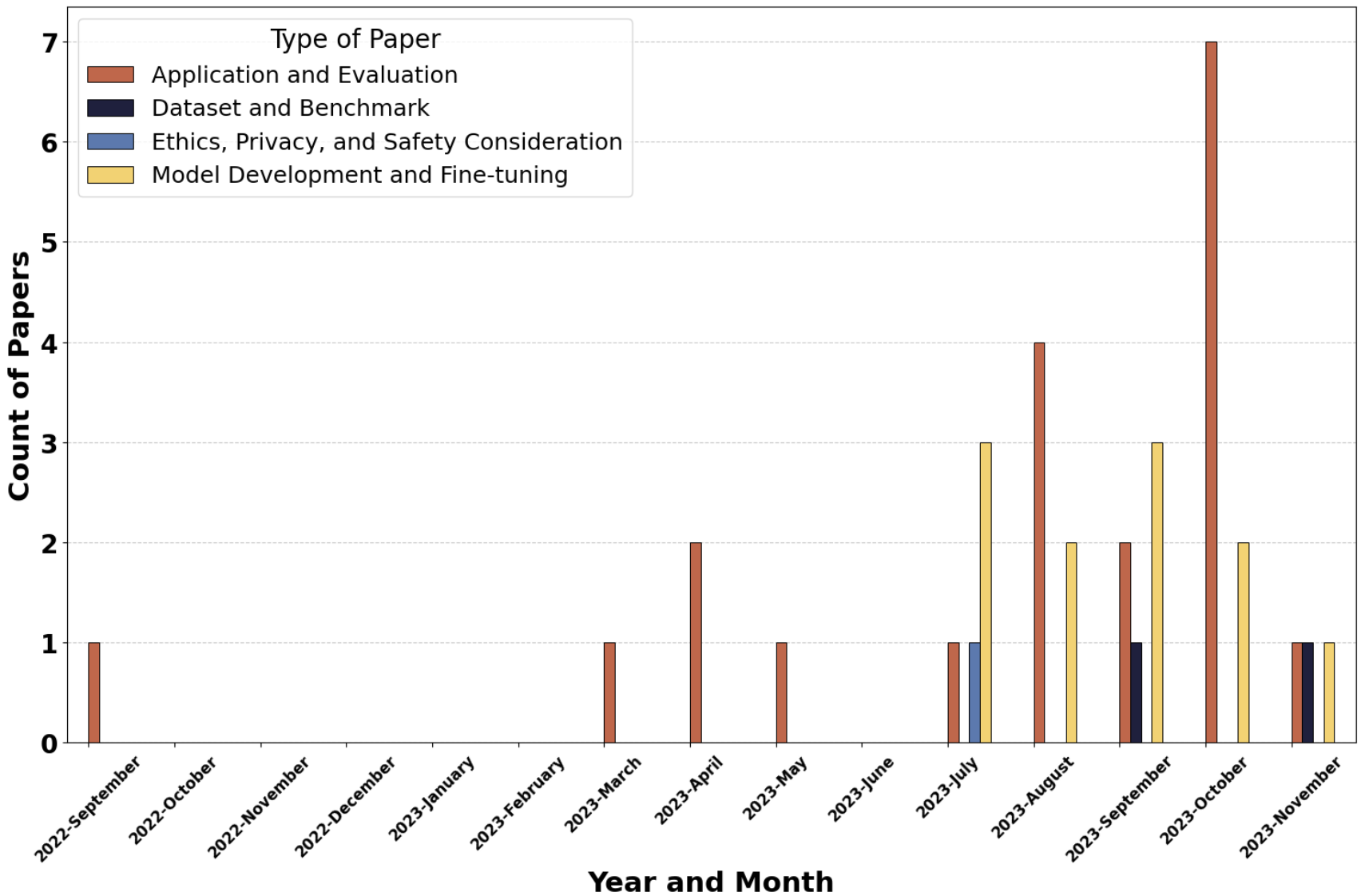}
    \caption{Submission/Publication time and type distribution of qualified studies.}
    \label{fig:paper_dist}
\end{figure}

Figure \ref{fig:paper_dist} shows the submission/publication time\footnote{Papers with pre-prints have submission time entries.} and type of the papers included in the final analysis. Research on LLMs in mental health care emerged in Sep 2022 and has shown a progressive increase in publication volume, with a surge in the latter half of the year, particularly noticeable in October. The majority of these papers are concentrated in the "Prompt-tuning and Application" category, which saw a rise in July. In contrast, "Model Development and Fine-tuning" studies were absent in the initial months, only to experience a marked rise in October. Notably, there have been only two papers concerning "Dataset and Benchmark", both emerging later in the year. Additionally, there is only one study \cite{2} that addresses ethics, privacy, and other challenges, which was published mid-year. 

\subsection{Areas of Application and Mental Health Conditions Involved}
During our review, we observed a strong correlation between the scope of each study and the dataset employed. This section offers a high-level overview of the application areas across these studies, along with the mental health conditions they intended to address. In Table \ref{tab:data}, we present a comprehensive summary of all the datasets utilized by these studies, detailing their intended purposes.

\subsubsection{Application Areas}
The studies roughly cover three main areas. The first area is the deployment of conversational agents (also called "chatbots") \cite{1,2,8,10,12,13,14,16,19,21,22,25,29,46,47,50}, with 16 studies aiming to improve the models’ capacity to produce empathetic and context-aware responses. These agents are typically not tailored to specific mental disorders, instead, they cater to a broad spectrum of mental health needs. Among these studies, 12 are designed to interact directly with individuals seeking mental health support through various platforms such as personal digital companions \cite{2}, on-demand online counseling \cite{8,10,12,14,16,19,22,25,46,47,50}, and emotional support \cite{21}. Some studies extend to specialized applications like couple therapy \cite{46,47}. The remaining four studies are developed to assist care providers by providing specific recommendations and analyses\cite{10,13}. This helps to mitigate the issue of provider shortage while enhancing the delivery of care. It’s noteworthy that some studies have taken a different approach by developing screening tools. These tools are designed to identify urgent cases \cite{10} and distinguish between emotional experiences in individuals with different personality disorder subtypes\cite{29}.

The second area focuses on resource enrichment. Studies in this category \cite{6,9,26,45,49} have explored LLM-generated explanations for multi-task analysis \cite{6,9} and the development of educational content\cite{45, 49}. The latter includes creating hypothetical case vignettes relevant to social psychiatry \cite{45} and personalized psychoeducation material\cite{49}. The potential of synthetic data provided by LLMs is also being tapped into as a method for augmenting data and fine-tuning downstream, particularly in augmenting clinical questionnaires to enrich the symptomatology of depression cases\cite{26}.

The third area involves the use of LLMs as classification models for detailed diagnosis\cite{4,5,6,9,17,28,42}. This often involves binary classification to detect the presence or absence of a single condition providing given contexts\cite{5,9,28,42}, and multi-class classification for more granular information about a condition, including severity levels and subtypes\cite{4,5,6,9,17}. Section \ref{conditions} provides more information about the mental health conditions explored in these studies. Some examples of multi-class classification include depression severity prediction (minimal, mild, moderate, and severe according to DSM-5 \cite{american1994diagnostic}), subtypes of suicide (supportive, indicator, ideation, behavior, and attempt according to the Columbia Suicide Severity Rating Scale (C-SSRS) \cite{posner2011columbia}), and identification of sources of stress span (school, finance, family, social relation, etc., based on the classifications established in SAD\cite{mauriello2021sad}.

\subsubsection{Mental Health Problems Involved}\label{conditions}
Out of the 34 articles reviewed, 23 focused on specific mental health problems, while the remaining articles explored general mental health knowledge or dialogues without specific conditions. Of the 23 articles examining specific mental health problems, 14 focused on single mental health problems, and nine involved multiple.

A wide range of mental health conditions has been studied, but most of them are limited by the availability of opensource datasets. The most frequently studied conditions include stress\cite{4,5,6,7,9,10,12,21,47}, suicide\cite{4,5,6,9,13,28,42}, and depression\cite{4,5,6,9,11,14,24,26,31,39}. Other conditions include anxiety\cite{9,14}, bipolar disorder\cite{9}, Post-traumatic demoralization syndrome (PTDS)\cite{9,27}, Autism spectrum disorder (ASD)\cite{22}, and personality disorder\cite{29}, loneliness\cite{9,28}, cognitive distortion\cite{20}, In addition to these conditions, life circumstances that can profoundly affect mental health are also studied, covering relationship problems\cite{28,46,47}, sleep problems, sexual violence, custody issues, and bullying\cite{28}.

\begin{table*}[t]
\scriptsize
\centering
\setlength{\tabcolsep}{5pt}
\caption{Summary of existing LLMs for mental health, in terms of their model development, the number of parameters, the availability of training data, the training strategy, and the available public links. "B" denotes billion. {\tfp} and {\ift} stand for "tuning-free prompting" and "instruction fine-tuning", respectively. A Base Model is a foundational model trained on a general dataset, which serves as the starting point for fine-tuning. Therefore, in the "Base Model" column, a model like ChatGPT would list its foundation, which is GPT-3.5.}
\resizebox{\textwidth}{!}{
\begin{tabular}{llcccc}
\toprule
 \multirow{1}{*}{Method}  & Base LLM & \multirow{1}{*}{\# Params} & Base Model Training Data & \begin{tabular}[c]{@{}c@{}} Training  Strategy \end{tabular} & \multirow{1}{*}{\begin{tabular}[c]{@{}l@{}} Availablity  \end{tabular}} \\
\midrule
DSC \cite{1} &  GPT-3.5\cite{gpt35}  & \unknown & \unknown & \tfp &  \xmark  \\
Replika \cite{2} & GPT-3\cite{gpt3}  & 175B  & Partially known & \tfp&  \cmark     \\  
Mental-LLM \cite{4} & Alpaca\cite{alpaca}/FLAN-T5\cite{t5}  & 7B/11B  &Partially known/Fully known &  \ift    &  \cmark  \\  
MentaLLaMA \cite{6,9} & LLaMA-2\cite{llama2}  & 33B   & Partially known & \ift  &  \cmark \\  
MindShift \cite{7} & GPT-3.5  & \unknown   & \unknown & \tfp  &    \xmark    \\  
Psy-LLM \cite{10} & Pangu\cite{zeng2021pangualphalargescaleautoregressivepretrained}   & 350M   & Partially known & \tfp  &    \xmark    \\  
MindfulDiary \cite{11} & GPT-4\cite{gpt-4}  & \unknown & \unknown & \tfp   &    \xmark    \\ 
ChatCounselor \cite{12} & Vicuna\cite{vicuna}  & 7B   &Partially knwon &  \ift  &   \cmark   \\
LLM-Counselors \cite{13} & GPT-3.5  & \unknown   & \unknown & \tfp  &    \xmark      \\ 
Task-Adaptive Tokenization \cite{16} & LLaMA  & 7B   & Partially known & \tfp  &    \cmark      \\
Chain of Empathy Prompting \cite{19} & GPT-3.5   & \unknown & \unknown & \tfp    &   \xmark      \\  
  Diagnosis of Thought Prompting \cite{20} & GPT-3.5/GPT-4   & \unknown/\unknown     &  \unknown  & \tfp &  \cmark   \\  
 \citet{21} & LLaMA\cite{llama}  & 7B   & Public &  \ift  &  \cmark      \\  
\citet{24} & Vicuna/GPT-3.5/LLaMA/Mistral\cite{jiang2023mistral}   & 13B/\unknown/33B/7B     &  \unknown   & \tfp/\ift &  \xmark  \\  
 BBMHR  \cite{25} & BlenderBot-BST \cite{blenderbotbst}  & 2.7B    &  Public &  \ift  &  \cmark      \\  
 ChatCBPTSD  \cite{27} & GPT-3.5  & \unknown  & \unknown & \tfp   &  \cmark       \\  
\citet{28} & FLAN-UL2\cite{tay2023ul2}   & 20B  &  Partially known & \tfp &  \xmark      \\  
\citet{29} & GPT-3.5   & \unknown  &  \unknown  & \tfp &  \xmark   \\  
\citet{31} & GPT-3  & \unknown     &  \unknown &  \tfp &   \xmark  \\  
\citet{39} & GPT-4  & \unknown   &  \unknown &  \tfp &   \xmark \\
 MindWatch \cite{42} & GPT-3.5 & \unknown  &  \unknown    &  \tfp &   \xmark  \\
\citet{45} &  GPT-3.5 & \unknown  &  \unknown    &  \tfp &   \xmark  \\
\citet{46} &  GPT-3.5 & \unknown  &  \unknown    &  \tfp &   \xmark  \\
\citet{47} & GPT-4  & \unknown      &  \unknown    &  \tfp &   \xmark  \\

\citet{49} & GPT-3 & \unknown    &  \unknown    &  \tfp &   \xmark  \\
\citet{50} & GPT-3 & \unknown    &  \unknown    &  \tfp &   \xmark  \\
\bottomrule 
\end{tabular}%
}
\label{tab:model}
\end{table*}

\subsection{Models and Training Techniques}
The suitability and effectiveness of a pre-trained model for mental health care are fundamentally influenced by its training data, size, and open-source status. These factors collectively determine the model's potential biases or lack of representativeness for certain tasks and populations. In Table~\ref{tab:model}, we provide a summary of existing LLMs developed for mental health care. This summary includes details about underlying base models, the scale indicated by the number of parameters, the transparency of the base model training data, the strategies employed during training, and information regarding accessibility if they are open-source.

\subsubsection{Distribution of Model Architectures and Prompting Algorithms}
A significant amount of existing work has been dedicated to directly prompting GPT-3.5 or GPT-4 for mental health applications without training. Some examples include depression detection\cite{31,24}, suicide detection\cite{28}, cognitive distortion detection\cite{20}, and relationship counseling\cite{46,47}. In these studies, LLMs function as intelligent chatbots, engaging with users to provide a range of mental health services, including analysis\cite{2,5,6}, prediction \cite{4} and support\cite{10,11,13}.

To enhance their effectiveness, methods like few-shot prompting, which presents the LLMs with a small number of task demonstrations before requiring them to perform a task, and chain-of-thought (CoT) prompting\cite{wei2023chainofthought}, which prompt the model to generate intermediate steps or paths of reasoning when dealing with problems, are employed. Drawing from the success of CoT in NLP, novel approaches such as chain-of-empathy prompting\cite{19}, which incorporates the insights from psychotherapy (i.e., the therapists’ reasoning process) to prompt the LLMs to generate the cognitive reasoning of human emotion, and diagnosis-of-thought prompting\cite{20}, which prompt the LLMs to make a decision using three diagnosis stages (i.e., subjectivity assessment, contrastive reasoning, and schema analysis), have been proposed to further improve LLM performance in the mental health domain.


Another stream of research focuses on the further training or fine-tuning of general LLMs using mental health specific texts. This approach aims to inject mental health knowledge into existing base LLMs, leading to more relevant and accurate analyses and support. Notable examples include MentaLLaMA \cite{6,9} and Mental-LLM\cite{4}, which fine-tuned the LLaMA-2 model and the Alpaca\cite{alpaca}/FLAN-T5\cite{flan-t5} model, respectively, using social media data for enhanced mental health predictions. Similarly, ChatCounselor \cite{12} leveraged the Psych8k dataset\cite{12}, comprising real interactions between clients and psychologists, to fine-tune the Vicuna\cite{vicuna} model. Additionally, \citet{21} employed the LLaMA model, fine-tuned on emotional support dialogues.

While some studies used open-source LLMs that provide more transparency about their sizes and training data sources, the complete specifics of any base model’s training data remain undisclosed. For example, although overarching details about the training data for LLaMA and T5 are available, the detailed documents or sources used in these datasets are not publicly shared.

\subsubsection{Fine-Tuning Techniques}
Acknowledging the high cost and extensive time required to train LLMs from scratch\cite{zhou2023survey, zeng2022greenplm,ye2023qilin,guo2023continuous}, existing studies have uniformly adopted fine-tuning existing models on mental health data. This approach allows the models to acquire expert domain knowledge and evolve into dedicated mental health-focused LLMs.

All the studies that involved fine-tuning adopted the instruction fine-tuning (IFT) technique. IFT is a type of fine-tuning where a model is fine-tuned for an instruction-following task that involves instructing a model to perform another task. In contrast, in classical types of supervised fine-tuning, the model does not have access to instructions but is directly tuned to perform a single downstream task. IFT allows domain knowledge to be injected into LLMs while improving LLM’s capability to follow human instructions. For example, giving a conversation between clients and psychologist, ChatCounselor \cite{12} used the prompts: "\textit{Your task is to identify the patient and counsellor in the conversation. Summarize the conversation into only one round of conversation, one query or description by the patient and one feedback by the counsellor}", to prompt GPT-4 \cite{gpt-4} to generate instruction-input-output triples, in which the instruction could be "\textit{If you are a counsellor, please answer the questions based on the description of the patient}." The input could be the query/question from the patient and output could be the feedback/answer from the model.


\begin{table*}[!ht]
\centering
\caption{Overview of Mental Health-Related Datasets. \textsuperscript{+}Dataset curated or compiled in "Dataset and Benchmark" studies. "K" denotes a thousand. Information not provided is marked by {\np}. {\na} stands for not applicable.}
\setlength{\tabcolsep}{2pt}
\resizebox{\textwidth}{!}{
\begin{threeparttable}
\renewcommand{\arraystretch}{1.3}
\begin{tabular}{lm{4cm}m{3cm}cccccc}
\toprule
 Name  & Intended Task & Mental Condition & Data Source &  Sample size &  Constructed by & Reviewer experience & License \\
\midrule
\multicolumn{8}{c}{\textbf{Published Dataset}}\\
\midrule

PsyQA \cite{sun-etal-2021-psyqa} &  Counseling & \na & Mental health service platforms & 22,346 questions with 56,063 answers & Human & Trained laymen & \np  \\

SMHD \cite{cohan-etal-2018-smhd} &  Mental health condition detection& ADHD\tnote{1}, anxiety disorders, ASD\tnote{2}, bipolar disorders, BPD\tnote{3}, depression, eating disorders, OCD\tnote{4}, PTSD\tnote{5}, schizophrenia, SAD\tnote{6}
 & Reddit & 4,399K posts & Weak supervision & \np & CC-BY-4.0  \\

D4 \cite{yao2022d4} &  Depression diagnosis-oriented dialogues & depression & Human-machine dialogue & 1,339 dialogues & Human & Experts & Research only  \\

Esconv \cite{esconv} &  Emotional support conversation & \na & Crowdworker chatting simulation & 1,053 dialogues (31,410 utterances) & Human & Trained laymen & CC-BY-NC-4.0  \\

DialogueSafety \cite{qiu2023benchmark} &  Dialogue safety & \na & Online counseling platforms & 7,935 multi-turn dialogues & Human & Experts & MIT  \\

Dreaddit \cite{turcan-mckeown-2019-dreaddit} &  Stress detection &N/A& Reddit & 3,551 post segments & Human & Trained laymen & Publicly Available  \\

DepSeverity \cite{naseem2022early} &  Depression severity prediction & Depression & Reddit & 3,551 posts & Human & 
\np& Publicly Available  \\

\citet{haque2021deep} &  Suicide ideation detection & \na & Reddit & 1,895 posts & Weak supervision & \np & \np  \\

CSSRS-Suicide \cite{gaur2019knowledge} &  Suicide risk prediction &\na& Reddit & 500 users & Human & Experts & CC-BY 4.0  \\

Depression\_Reddit \cite{pirina-coltekin-2018-identifying} &  Mental health condition detection & Depression& Reddit & 800 posts & Weak supervision & \np & Publicly available  \\

CLPsych15 \cite{coppersmith-etal-2015-clpsych} &  Mental health condition detection & Depression \& PTSD & Twitter & 1,746 users & Human & \np & Research only  \\

\citet{ji2022suicidal} &  Suicide ideation \& mental disorder detection & Depression \& PTSD & Twitter & 866 users \& 59,212 posts & Weak supervision & \np & Research only  \\

SAD \cite{mauriello2021sad} &  Stress cause detection & \na & SMS messages & 6,847 SMS-like sentences & Human & Laymen & CC-BY-4.0  \\

CAMS \cite{garg-etal-2022-cams} &  Depression \& suicide cause detection & Depression & Reddit & 5,051 posts & Human & Trained students & Publicly available  \\

EMPATHETICDIALOGUES \cite{rashkin-etal-2019-towards} &  Empathetic dialogue generation & \na & Crowdworker chatting simulation & $\sim$25K multi-turn dialogues & Human & Laymen & CC-BY-NC-4.0  \\

MultiWD \cite{sathvik2023multiwd} &  Wellness dimensions detection & \na & Reddit & 3,227 posts & Human & Trained students \& experts & Research only  \\

IRF \cite{garg-etal-2023-annotated} &  Interpersonal risk factors detection & \na &Reddit & 3,523 cases & Human & Trained students & Research only  \\

GoEmotions \cite{demszky2020goemotions} &  Emotion classification & \na &Reddit & 58,009 cases & Human & Laymen & Apache 2.0  \\

MHP Reddit\cite{lahnala-etal-2021-exploring} &  Counseling &\na & Reddit & 9,501 question-answer pairs& Human & \np & MIT  \\

EPITOME \cite{sharma2020computational} & Empathetic response generation & \na &Reddit, TalkLife\cite{talklife} & 10K interactions on empathy & Human & Trained laymen & Publicly available  \\

Therapist Q\&A \cite{shreevastava2021detecting} & Cognitive distortions detection & \na & Kaggle & 2,531 patient speech samples & Human & \np & \np  \\

ExTES \cite{21} & Emotional support dialogue & \na &  ChatGPT &  11,177 dialogues & LLM generation & Trained laymen & Publicly available  \\

BDI-Sen \cite{perez2023bdi} & Mental health condition detection & Depression & Reddit & 4,973 sentences & Human & Experts & Research only  \\

PsySym \cite{zhang2022symptom} & Mental health condition detection & Depression, anxiety, ADHD, bipolar disorder, OCD, PTSD, eating disorder & Reddit &  83,779 sentences & Human & Trained laymen \& experts & \np  \\

NVDRS\cite{NVDRS} & Suicide behavior detection & \na &  Mortality \& incident reports &  1,462 reports & Human & Experts & \np  \\

EATD-Corpus \cite{shen2022automatic} & Mental health condition detection & Depression & Interview transcripts & 162 interview audio \& transcripts & Weak supervision & \np & Publicly available  \\

ANEW\cite{bradley1999affective} & Lexicon sentiment detection  & \na & English vocabulary &1K words & Human & \np & \np  \\

\citet{warriner2013norms} & Lexicon sentiment detection &N/A& English vocabulary &  13,915 lemmas & Human & Trained laymen & \np  \\

NRC-VAD \cite{mohammad-2018-obtaining} & Lexicon sentiment detection &N/A& English vocabulary &  20K words & Human & Laymen & \np  \\

\midrule

\multicolumn{7}{c}{\textbf{Self-constructed Dataset}}\\
\midrule

Ma et al.\cite{2} & User comment analysis & \na &Reddit &  2,917 posts & Human & \np & \np  \\

Psych8k\cite{12} & Mental health support & \na &Real-life counseling &  260 interviews with 8,187 instruction pairs & \np & \np & Publicly available  \\

Yao et al. \cite{14} & Counseling & Postpartum Mood \& Anxiety Disorders & Helpline &  7,014 
conversations with 65,062 messages & Weak supervision & \np & \np  \\

Qi et al. \cite{17}$^{+}$ & Suicide risk prediction \& cognitive distortion detection & N/A & Weibo & 2,159 comments & Human & Experts & Publicly available  \\

Bhaumik et al.\cite{42} & Suicide ideation detection \& personalized psychoeducation &\na& Reddit &  232K posts & \np & \np & \np  \\

Kumar et al.\cite{50} & Counseling &N/A& Crowdworker chatting simulation &  945 survey responses & Human & Laymen & \np  \\

\bottomrule
\end{tabular}
\begin{tablenotes} %
\item[1] Attention-deficit/hyperactivity disorder.
\item[2] Autism spectrum disorders
\item[3] Borderline personality disorder.
\item[4] Obsessive-compulsive disorder.
\item[5] Post-traumatic stress disorder.
\item[6] Seasonal affective disorder.
\end{tablenotes}
\end{threeparttable}

}%
\label{tab:data}
\end{table*}


\subsection{Dataset Characteristics}
The integrity of the data profoundly influences outcomes in mental health care studies, particularly in terms of representativeness, quality, and potential bias. Recognizing the sources and characteristics of datasets is fundamental in ensuring accurate and fair research findings. We reviewed datasets used in the studies and reported their associated tasks, data sources, sample sizes, annotation methods if the dataset is annotated, human reviewer experience, and licenses in Table \ref{tab:data}. 

Among the 34 studies we reviewed, 36 datasets were identified. These datasets contain diverse data applicable to a wide range of intended mental health care tasks. Most datasets are dedicated to detection or classification tasks. This category includes tasks such as mental health condition detection, mainly focusing on depression and post-traumatic stress disorder (PTSD) \cite{naseem2022early,pirina-coltekin-2018-identifying,coppersmith-etal-2015-clpsych,turcan-mckeown-2019-dreaddit,zhang2022symptom}, cause of stress detection \cite{mauriello2021sad}, and interpersonal risk factor prediction \cite{sathvik2023multiwd}. Another major group focuses on text-generation tasks, such as simulating counseling sessions\cite{sun-etal-2021-psyqa,lahnala-etal-2021-exploring} responding to medical queries and generating empathetic dialogue\cite{rashkin-etal-2019-towards,sharma2020computational}. The remainder concerns more specialized applications, such as analysis of user discussions of emotion-support LLMs \cite{ma2023understanding} and dialogue safety exploration\cite{qiu2023benchmark}.

Sources of these datasets are imbalanced, with 20 datasets (55.6\%) collected from social media platforms, such as Reddit, Twitter, and Weibo. Eleven datasets relied on more controlled venues, such as mental health counseling or interview dialogues simulated by clinicians \cite{qiu2023benchmark} and collected from real-life scenarios\cite{sun-etal-2021-psyqa,yao2022d4, 12}. Suicide behavior detection corpus collected from mortality and incident reports\cite{NVDRS}, and helpline conversations\cite{14}. Less frequently seen sources include LLM-synthesized data \cite{21}, existing sentiment lexicons\cite{bradley1999affective,warriner2013norms,mohammad-2018-obtaining}, and crowdworker simulated conversation\cite{esconv,rashkin-etal-2019-towards,50}.

Dataset sizes and units vary with sources and annotation methods. Those drawn from social media and created via web crawling with the need for annotation typically have larger sizes. In contrast, those composed of expert content contain fewer samples. For example, SMHD \cite{cohan-etal-2018-smhd} contains 4,399K posts collected from Reddit, while D4 \cite{yao2022d4} only contains 1,339 dialogues reviewed by experts \cite{van2023if}.

Most datasets involved human review of the content regardless of their construction methods. Eight datasets explicitly mentioned the involvement of experts in the review process. Fourteen datasets did not specify the experience level of the annotators. The remaining datasets were reviewed by laymen or trained laymen, including those who might be crowdworkers such as Amazon Mechanical Turk (MTurk) workers. Most papers relied on publicly available datasets. Six datasets \cite{2, 12, 14,17,42} were self-constructed. Only one of them\cite{14}, curated in a "Dataset and Benchmark" study, has been released for public use. Most of these datasets are released under licenses that restrict their use to non-commercial purposes. Details and implications of these restrictions and licensing agreements are provided in Appendix \ref{sec:license}.

\subsection{Validation measures and metrics}
To ensure robust and unbiased assessments of LLMs,  it's crucial to adopt appropriate validation measures and metrics. Our analysis categorized them into two types: automated evaluation and human assessment. Table \ref{table:metrics} details metrics used in automated evaluation and attributes evaluated by human assessment. To reflect the appropriateness of these metrics, we also categorized them based on their evaluation focus: language proficiency and mental health applicability.

For mental health applicability, different variants of F1 scores are the most predominant metric adopted across studies \cite{5, 6, 9, 14, 19, 20, 24, 27, 31}. Accuracy, as another fundamental metric, is also extensively employed \cite{3, 5, 6, 9, 13, 20, 42}. Recall (also known as sensitivity) \cite{4, 6, 14, 19, 27, 42} and precision \cite{14, 19, 24, 26, 42} are also widely used, and frequently reported with F1 and accuracy. Diagnosis-focused studies, while also heavily relying on F1 and Accuracy, tend to employ additional metrics such as Area Under the Receiver Operating Characteristic (AUROC)\cite{27, 42}, and Specificity \cite{27} for a holistic understanding of LLM diagnostic validity. 

In evaluating language proficiency, BLEU\cite{1, 3, 6, 16, 21, 24, 25}, ROUGE\cite{6, 10, 16, 21, 24}, Distinct-N\cite{1, 3, 10, 21}, and METEOR \cite{21} are widely used to assess alignment with human-like language, diversity of expression, and overall quality of the generated text. Advanced metrics such as GPT3-Score\cite{6}, BARTScore\cite{6}, and BERT-Score \cite{14, 16, 25} are tailored to evaluate semantic coherence and relevance of text in specific contexts, addressing the demand for nuanced understanding. Perplexity \cite{10} quantifies model predictability and text naturalness, while Extrema and Vector Extrema \cite{21} analyze the diversity and distribution in the embedding space to reflect the model's linguistic creativity and depth. The adoption of these traditional language evaluation metrics appears to be driven by a lack of automated, efficient, and easily understandable metrics designed for evaluating the quality of free-text generation of LLMs for mental health care. Human evaluation is thus frequently used across studies. 

Among the 34 articles reviewed, 19 employed a combination of human and automated evaluations, 5 exclusively used human evaluations, and 10 relied solely on automated methods. However, there is no single unified evaluation framework. A few studies \cite{1, 8, 19, 21, 25} adopted or adapted published evaluation standards or attributes discussed in previous studies\cite{sharma2020computational, liu2021towards, o2018suddenly, li2019acuteeval, qiu2023benchmark}, yet none of these frameworks are widely adopted. Some of these studies and most other studies added or designed from scratch their evaluation frameworks and selected attributes from interest from their own judgment\cite{9,10,12,13,16,21, 22, 25}. A wide range of properties/attributes are proposed to be evaluated, and the frequently overlapped ones are empathy\cite{1, 14, 22, 25, 47}, relevance\cite{1, 3, 10, 12, 24}, fluency\cite{1, 3, 6, 10, 16, 21}, understanding\cite{21, 22}, and helpfulness\cite{10, 21, 25, 47}. Specifically, in intervention applications, studies often measure user engagement with interventions and technology adoption. The rate of intervention acceptance\cite{7, 49, 50}, frequency of smartphone usage\cite{7}, adherence to digital journaling protocols\cite{11}, willingness to use the intervention again\cite{47,49,50}, and intervention adherence rates \cite{7,11,47,49,50} are observed to evaluate the effectiveness of LLMs in these studies.

We noticed some attributes, despite sharing names, differ in definition across studies. For instance, 'Informativeness' varies in context – in some studies\cite{1,12} (\cite{1} did not directly use informativeness, so it is not included in table \ref{table:metrics}), it relates to the richness of LLM responses, while in others\cite{21}, it measures the extent of individual elaboration on emotional distress. Expert-led assessments \cite{13, 22, 39,45, 47} prominently involve direct analysis of model outputs or expert questionnaire scoring. Reliability metrics like Cohen’s Weighted Kappa\cite{1, 14}, Krippendorff’s Alpha\cite{13}, Fleiss’ Kappa\cite{6, 21}, and inter-annotator agreement \cite{25, 46} have been essential for validating study instruments. The study designs vary, with the number of reviewers ranging from as few as three \cite{24} to as many as fifty\cite{21}.

\begin{table*}[ht!]
\centering
\caption{Automated Evaluation Metrics and Human Evaluation Metrics Used in More Than One Study. Metrics are ordered alphabetically for easy indexing. Only model-oriented metrics are shown for simplicity.}
\setlength{\tabcolsep}{2pt}
\resizebox{\textwidth}{!}{
\begin{tabular}{lcm{7cm}c} 
\hline
\multicolumn{4}{c}{\textbf{Automated Metrics}} \\
\hline
\textbf{Metric} & \textbf{Associated studies} & \textbf{Metric explanation} & \textbf{Target of Evaluation} \\ 

Accuracy & \cite{3, 4, 5, 19, 31, 42} & Proportion of correct predictions in classification tasks. & \mentalhealth \\

AUROC & \cite{27, 42} & Measures the model's performance in distinguishing between classes. & \mentalhealth \\

BERT-Score \cite{zhang2020bertscore}/BARTScore \cite{bartscore}/GPT3-Score\cite{gptscore}  & \cite{6, 9, 14, 16, 25} & Evaluate the quality of text generation against reference texts using respective models. & \lang \\

BLEU-N\cite{bleu} & \cite{1, 3, 6, 16, 21, 24, 25} & Measures the overlap of n-grams between the generated text and reference texts. & \lang \\

Distinct-N & \cite{1, 10, 21}  & Measures the diversity of language by calculating unique n-grams. & \lang \\

F1 Score & \cite{5, 6, 9, 14, 17,  19, 20, 21, 24, 27, 28, 31} & A combination of precision and recall into a single metric, balancing the two. & \mentalhealth \\

Precision & \cite{14, 17, 19, 26, 42} & The proportion of positive identifications that were actually correct. & \mentalhealth \\

Recall (sensitivity) & \cite{6, 14, 17, 19, 27, 42} & The proportion of actual positives that were identified correctly. & \mentalhealth \\

ROUGE-N/L \cite{lin-2004-rouge}& \cite{6, 10, 16, 21, 24, 25} & Measures the overlap of n-grams or longest common subsequence between texts. & \lang \\

Specificity & \cite{25, 27} & The proportion of true negatives that are correctly identified. & \mentalhealth \\
\hline
\multicolumn{4}{c}{\textbf{Attributes Evaluated in Human Assessment}} \\
\hline
\textbf{Attribute} & \textbf{Associated studies} & \textbf{Attribute definition} & \textbf{Target of Evaluation} \\ 

Coherence & \cite{3, 16, 21} & The chatbot's ability to keep the conversation focused and transition smoothly between topics. & \lang \\

Consistency & \cite{9, 21} & Ensuring the chatbot's responses are aligned with its defined role without contradictions. & \lang \\

Empathy & \cite{1, 14, 22, 25, 47} & The degree to which the chatbot shows warmth, sympathy, and concern towards the user's situation. & \mentalhealth \\

Engagement & \cite{11,22,25} & The chatbot's ability to maintain prolonged and engaging conversations with the user. & \mentalhealth \\

Exploration & \cite{3, 8, 11, 19} & Indicators of the user's attempts to investigate deeper into the topics brought up in the conversation. & \mentalhealth \\

Fluency & \cite{1, 3, 6, 10, 16, 21} & The chatbot's generation is grammatically correct, well-formed, and natural-sounding, resembling human-like speech or writing. & \lang \\

Guidance & \cite{12, 13} & The chatbot’s effectiveness in giving clear, actionable instructions and guidance. & \mentalhealth \\

Helpfulness & \cite{10, 21, 25, 47} & The extent to which the chatbot contributes to reducing the user's emotional distress and improving mood. & \mentalhealth \\

Informativeness & \cite{12, 21} & The extent to which the chatbot provides useful and relevant suggestions for the user's issues. & \mentalhealth \\

Interpretation & \cite{3, 7, 12, 13, 19} & The user's effort to restate or reinterpret the problems presented, as indicated by linguistic markers. & \mentalhealth \\

Professionalism & \cite{9, 13, 16} & Evaluating the chatbot's responses for their logical and psychological accuracy. & \mentalhealth \\

Relevance & \cite{1, 3, 10, 12, 24} & The chatbot's ability to generate content related to the user's specific problems. & \mentalhealth \\

Understanding & \cite{21, 22} & The chatbot's level of accurately grasping and reflecting the user's experiences and emotions. & \mentalhealth \\

Reliability & \cite{6,9} & The chatbot's generation is factually accurate, logically consistent, and dependable across similar inputs or contexts. & \mentalhealth \\

\hline

\end{tabular}}
\label{table:metrics}
\end{table*}

\subsection{Benchmark studies}
Benchmark studies are included in this application-focused review to identify areas needing enhancement for objectively assessing the efficacy of various LLMs in mental health care. So far there have been two benchmark studies\cite{3,17}. Jin et al. \cite{3} provides a comprehensive evaluation across models including GPT-4, GPT-3.5, Alpaca\cite{alpaca}, Vicuna\cite{vicuna}, and LLaMA-2\cite{llama2}, on six tasks using data from social media and therapy sessions: diagnosis prediction, sentiment analysis for inferring emotional states, language modeling for understanding conversations, and question answering for therapeutic guidance, along with evaluating therapy session dialogues. In contrast, Qi et al. \cite{17} have a much more focused scope on classifying cognitive distortions and predicting
suicidal risks on Chinese social media data. They evaluated ChatGLM2-6B\cite{chatglm}, GLM-130B\cite{glm130b}, GPT-3.5, and GPT-4.

\section{Discussion}\label{sec:discussion}
Through an examination of 34 pertinent studies, we synthesized the characteristics, methodologies, datasets, validation measures, application areas, and specific mental health challenges addressed by LLMs. This synthesis aims to serve as a bridge between the computational and mental health communities, detailing insights in an accessible manner. Despite preliminary evidence suggesting the feasibility of LLMs in enhancing mental health care, our results show that notable gaps remain between their current state and actual clinical applicability. Below, we outline three key areas for improvement that could help realize the full potential of LLMs in clinical practice.

\textbf{Improvements in data quality for LLM development and validation: }The predominant method in mental health LLM applications is prompt-tuning favored for its simplicity and conversational interaction style. GPT-3.5 and GPT-4, widely recognized as ChatGPT, are at the forefront of this research. Despite their impressive performance, they occasionally underperform in the complex context of mental health and may yield biased results. As GPT-4 is not modifiable, it is necessary to investigate fine-tuning techniques on open-source LLMs to mitigate these issues.

Current fine-tuning efforts mainly utilize dialogue and social media data, like Reddit postings. While this data can be useful for screening, it falls short of establishing formal clinical diagnoses due to oversimplifications and the diverse nature of clinical symptoms\cite{harrigian-etal-2020-models,hua2022using}. In addition, they were sourced from a distribution of users that very likely do not represent the general population at risk of the mental conditions in question, and descriptions of compositions of the included users are lacking. Moreover, labels in existing datasets are typically not curated or validated by mental health experts, with substantial variability in their definitions across studies. For example, "depression" and "anxiety" range from everyday mood fluctuations to clinical disorders. Only one study \cite{4} made a distinction between clinical diagnoses and self-reported mental states. For LLMs to be reliably integrated into clinical workflows, development and validation with high-quality data are crucial. This data should accurately define mental health conditions and provide a solid basis for rigorous assessment.


\textbf{Enhanced reasoning, empathy capabilities, and evaluation methods:} 
Effective dialogue-based tasks, especially those on mental health care, require advanced reasoning and empathy abilities, allowing for the analysis of statements provided by the users, and the provision of appropriate feedback. However, the extent to which LLMs’ reasoning is a product of their large training datasets or a genuine emergent property remains a subject of debate (i.e., memorization vs. understanding/reasoning). In cases where "reasoning" is based on memorized training data, the models may lack robustness, particularly in handling diverse user dialogues and language use. In addition, although LLMs have shown preliminary capabilities in demonstrating empathy, these results are still inconclusive.

The absence of a unified evaluation framework amplifies this issue. While discriminative tasks with clear right or wrong answers are easy to assess, evaluating free-text generation is challenging due to the absence of a universal standard. Previous studies have used automated metrics like BLEU, ROUGE-L, and Meteor to evaluate language quality, but these fail to capture context. Even advanced metrics like the BERT score and GPT-3 score fall short in capturing the nuances of mental health care dialogues. Most studies have resorted to human evaluations, but the lack of a standardized framework leads to subjective selection of assessment aspects, resulting in varied and potentially incomparable results. This methodology discrepancy hinders the generalization of findings and underscores the need for a more relevant evaluation framework. 

\textbf{Addressing privacy, safety, and ethical/regulatory concerns:} Prioritizing patient privacy, safety, and ethical standards is paramount in mental health applications. For LLMs to serve as reliable clinical aids that can be applied in practice, models are expected to process sensitive personal and health information; therefore, strict adherence to data protection laws like the Health Insurance Portability and Accountability Act (HIPAA) \cite{hipaa1996} and the General Data Protection Regulation (GDPR) \cite{GDPR2016a} is necessary. Standards for bias analysis, model transparency, and informed consent should be established and enforced. To maintain alignment with ethical standards and clinical efficacy, developers, mental health professionals, ethicists, and legal experts must collaborate closely and implement continuous monitoring and evaluation of the LLMs. Additionally, emergency protocols should be in place to allow for expert intervention when necessary.



\textbf{Continuing Trends in 2024:}
The ongoing scarcity of robust clinical datasets in mental health research likely sustains heavy reliance on social media-derived data (e.g., Reddit/X) as primary sources for model development in 2024. This trend mirrors our documented observations of their pervasive use in 2023 for training and validating mental health-related AI tools, raising critical questions about inherent biases, representativeness, and clinical generalizability.

The absence of standardized validation frameworks persists as a key methodological challenge. Our analysis identified marked heterogeneity in evaluation protocols during 2023, with no consensus emerging on performance metrics or clinical benchmarking criteria. This inconsistency will likely have extended into 2024, underscoring an enduring barrier to rigorous assessment of LLM efficacy, safety, and translational potential in mental healthcare contexts.

Furthermore, accelerating interdisciplinary integration between AI developers, clinical practitioners, and public health stakeholders, a trend catalyzed in 2023 by proliferating open-source LLM ecosystems, has probably intensified in 2024. Such collaborations hold promise for advancing context-sensitive model architectures, yet they concurrently necessitate robust governance frameworks to address ethical risks and ensure equitable implementation.

\section{Conclusion}
LLMs hold promise for enhancing mental health care with their ability to process and generate large volumes of textual information. Our results suggest that while it is feasible to run these models, concerns about the lack of valid training sets, lack of standardized evaluation metrics, and privacy/ethical concerns require careful consideration before any clinical implementation. Addressing these challenges will be more productive than further pilot studies which show the feasibility but also require new collaborations to refine technology, develop standardized evaluations, and ensure ethical applications, aiming to realize LLMs’ full potential in supporting mental health care.

\section*{Key References}
\label{sec:key_references} 
This section highlights several key references published within the review period that exemplify or directly inform the major challenges and future directions outlined in the Discussion, particularly concerning data quality, model capabilities, evaluation, and ethical considerations.

\begin{itemize}
    \item \citet{4}: As noted in the discussion, this work is important for attempting to differentiate between clinical diagnoses and self-reported mental states, directly addressing the crucial challenge of data definition and quality for reliable LLM application in mental health.
    \item \citet{harrigian-etal-2020-models}: This study, cited in the discussion, critically examines the generalizability of mental health models trained on social media data, underscoring the need for improved, clinically relevant datasets emphasized in this review.
    \item \citet{19}: This paper exemplifies recent efforts to enhance LLM capabilities crucial for mental health care, specifically empathy. It directly addresses the need for improved model functionalities and nuanced interaction methods discussed as vital for progress, even if current results remain inconclusive.
    \item \citet{3}: Directly confronts the lack of standardized evaluation metrics identified as a major hindrance in the discussion. Proposing a dedicated benchmark for mental health LLMs represents a critical step towards comparable assessments of model performance and safety.
    \item \citet{9}: This study exemplifies the fine-tuning of open-source models (LLaMA 2) discussed as necessary to move beyond closed models like GPT-4. It also showcases work directly using social media data, highlighting both the potential and the challenges related to data quality and interpretation raised in the discussion.
    \item \citet{ma2023understanding}: This study provides user perspectives on LLM-based mental health support, directly informing the discussion around practical implementation challenges, including ethical considerations, safety, and the potential for over-reliance.
\end{itemize}


\bibliography{sn-article}

\begin{thebibliography}{100}
\bibitem[\protect\citeauthoryear{Khraisha et~al.}{2023}]{gpt4review2}
Khraisha Q, Put S, Kappenberg J, Warraitch A, Hadfield K.: Can large language models replace humans in the systematic review process? Evaluating GPT-4's efficacy in screening and extracting data from peer-reviewed and grey literature in multiple languages.
\end{thebibliography}

\begin{appendices}





\section{Statements and Declarations}
\subsection{Funding}
The authors declare that no funds, grants, or other support were received during the preparation of this manuscript.
\subsection{Competing Interests}
JT has research support from Otsuka and is an adviser to Precision Mental Wellness. All other authors have no conflict of interest.
\subsection{Author Contributions}
\textbf{Study design, data collection, and article screening}: Y.H.\\
\textbf{Result analysis}: Y.H., F.L., K.Y., and Z.L. \\
\textbf{Manuscript drafting}: Y.H., F.L., K.Y., Z.L., H.N., P.Z, and Y.S. \\
\textbf{Critical feedback}: Y.S., L.M., and A.B. \\
\textbf{Manuscript review}: All authors. \\
Y.H. takes the integral responsibility of the study.

\section{Supplementary}
\subsection{S.1. Screening prompt for ChatGPT-4}\label{sec:prompt_design}
\begin{figure}[th!]
    \centering
    \includegraphics[width=0.65\linewidth]{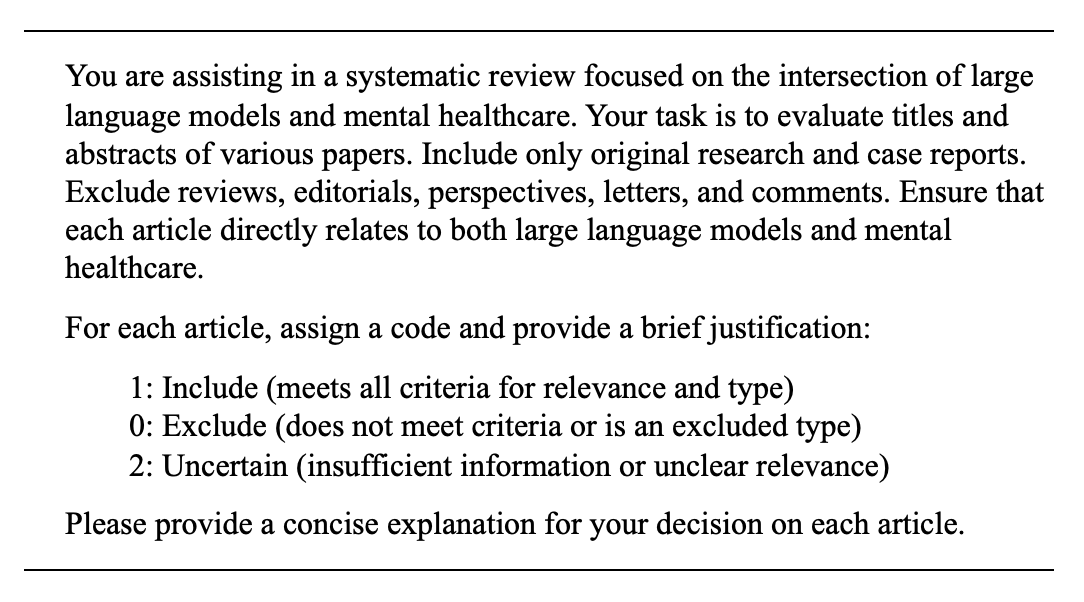}
    \caption{Screening prompt for ChatGPT-4.}
    \label{fig:prompt}
\end{figure}

The template shown in Figure~\ref{fig:prompt} was employed for screening titles and abstracts in the review's screening phase. After testing various prompts, the one presented here demonstrated the highest accuracy, achieving a perfect score (10 out of 10) in a randomly selected sample. It was crucial to specify that articles should be explicitly focused on LLMs and mental health care. This precision was necessary to avoid the inclusion of articles that were only vaguely related to the topic, a common issue encountered when the criteria were less specific.

\subsection*{S.2.: Data Restriction and Licensing in Related Datasets}\label{sec:license}
Dataset restrictions and licenses establish the guidelines for how data may be utilized, shared, altered, and disseminated. Users must comply with the terms set forth by the restrictions and licenses under which the datasets are released. Below is an explanation of the various restrictions and licenses associated with the datasets we reviewed:

\begin{enumerate}
    \item \textbf{Public Available}: Indicates that the dataset is made freely available to the public and can be used, distributed, or modified without any restrictions, sometimes even for commercial purposes, depending on the exact nature of the public domain status.
    \item \textbf{Research Only}: Indicates that the dataset is available only for non-commercial, academic research purposes. It typically restricts commercial use or widespread redistribution without explicit permission.
    \item \textbf{MIT License}: A permissive free software license originating at the Massachusetts Institute of Technology (MIT). It allows for almost unrestricted freedom with the work, including commercial use, with the only requirement being to preserve the copyright and license notices.
    \item \textbf{Apache 2.0}: A permissive free software license written by the Apache Software Foundation. It allows users to use the software for any purpose, to distribute it, to modify it, and to distribute modified versions of the software under the terms of the license, without concern for royalties.
    \item \textbf{CC-BY-4.0 (Creative Commons Attribution 4.0 International)}: Allows others to share, copy, distribute, execute, and transmit the work, as well as to remix, adapt, and build upon the work, even commercially, as long as they provide a credit to the original creator.
    \item \textbf{CC-BY-NC-4.0 (Creative Commons Attribution NonCommercial 4.0 International)}: Similar to CC-BY, this license allows others to remix, adapt, and build upon the work non-commercially, and although their new works must also acknowledge the creator and be non-commercial, they don’t have to license their derivative works on the same terms.
\end{enumerate}
Datasets that do not have a specific license mentioned or that the availability of the dataset is not explicitly stated are marked with {\np}. It is advisable to contact the dataset providers or creators for more information on usage rights and restrictions

\end{appendices}

\end{document}